\documentclass{article}

\usepackage{arxiv}

\usepackage[utf8]{inputenc} 
\usepackage[T1]{fontenc}    
\usepackage{hyperref}       
\usepackage{url}            
\usepackage{booktabs}       
\usepackage{amsfonts}       
\usepackage{nicefrac}       
\usepackage{microtype}      
\usepackage{xcolor}         
\usepackage{graphicx}
\usepackage{amsmath}
\usepackage{amssymb}
\usepackage{amsthm}
\usepackage{subcaption}
\usepackage{algorithm}
\usepackage{algorithmic}

\DeclareMathOperator*{\argmin}{arg\,min}

\newcommand{\E}{\mathbb{E}}
\newcommand{\what}[1]{\widehat{#1}}
\newcommand{\mbs}[1]{\boldsymbol{#1}}
\newcommand{\mbf}[1]{\mathbf{#1}}
\newcommand{\Prob}{\mathbb{P}}

\newcommand{\parm}{\mbs{\theta}}
\newcommand{\parmset}{\Theta}
\newcommand{\ctx}{c}
\newcommand{\y}{y}
\newcommand{\x}{x}

\newcommand{\zvec}{\mbs{z}}
\newcommand{\xvec}{\mbs{\x}}
\newcommand{\lossfunc}{\ell_{\parm}}
\newcommand{\eps}{\varepsilon}

\newcommand{\ndata}{n}

\newcommand{\ptrain}{p}

\newcommand{\ctxset}{\mathcal{C}}
\newcommand{\regret}{\Delta}
\newcommand{\prb}{p}

\newcommand{\prbset}{\mathcal{P}}
\newcommand{\typical}{\beta} 
\newcommand{\wt}{w}
 
\newcommand{\regrethat}{\what{\regret}}
\newcommand{\risk}{R}

\newcommand{\parmerm}{\widehat{\parm}_{\text{erm}}}
\newcommand{\parmmm}{\widehat{\parm}_{\text{min-max}}}
\newcommand{\expregret}{\Delta}
\newcommand{\expregrethat}{\widehat{\expregret}}
\newcommand{\parmdrr}{\what{\parm}_{\typical}}

\newcommand{\erm}{\textsc{erm}}

\newcommand{\f}{\widetilde{\expregret}}

\newcounter{result}
\newtheorem*{result}{\refstepcounter{result} Result~\theresult}

\newcounter{example}
\newtheorem*{example}{\refstepcounter{example} Example~\theexample}

\title{Robust Learning in Heterogeneous Contexts}

\author{Muhammad Osama \qquad Dave Zachariah \qquad Petre Stoica\\
  Department of Information Technology\\
  Uppsala University
}

\begin{document}

\maketitle

\begin{abstract}
We consider the problem of learning from training data obtained in different contexts, where the underlying context distribution is unknown and is estimated empirically. We develop a robust method that takes into account the uncertainty of the context distribution. Unlike the conventional and overly conservative minimax approach, we focus on excess risks and construct distribution sets with statistical coverage to achieve an appropriate trade-off between performance and robustness. The proposed method is computationally scalable and shown to interpolate between empirical risk minimization and minimax regret objectives. Using both real and synthetic data, we demonstrate its ability to provide robustness in worst-case scenarios without harming performance in the nominal scenario.
\end{abstract}

\section{Introduction}

Machine-learning methods often leverage large amounts of training data that is collected from several sources in varying contexts. For instance, image classification data is labeled by several people and the data itself may be captured in different environments with different backgrounds or levels of illumination \cite{arjovsky2019invariant}. More generally, data could be collected in contexts with different covariate distributions \cite{shimodaira2000improving,rojas2018invariant} or different interventions \cite{buhlmann2020invariance, magliacane2018domain}. 

In this paper, we consider a finite set of contexts, indexed by an integer $\ctx \in \ctxset$. In given context $\ctx$, training data is drawn independently and identically as
\begin{equation}
\zvec_i \sim \ptrain(\zvec|\ctx), \quad i=1 ,\dots, n_c,
\label{eq:data}
\end{equation}
where the training distribution is unknown. Let $\parm\in\parmset$ be a decision parameter in a specific task and $\lossfunc(\zvec)$ denote the loss at $\parm$ incurred for test point $\zvec$. Then each context has a conditional risk denoted
$$\risk_{\ctx}(\parm)=\E[\lossfunc(\zvec)|\ctx]$$
Using $n_c$ training samples the conditional risk can be approximated by an empirical average (or a regularized form of it).

\begin{figure*}[t!]
    \centering
    \begin{subfigure}{0.45\linewidth}
    \includegraphics[width=0.95\linewidth]{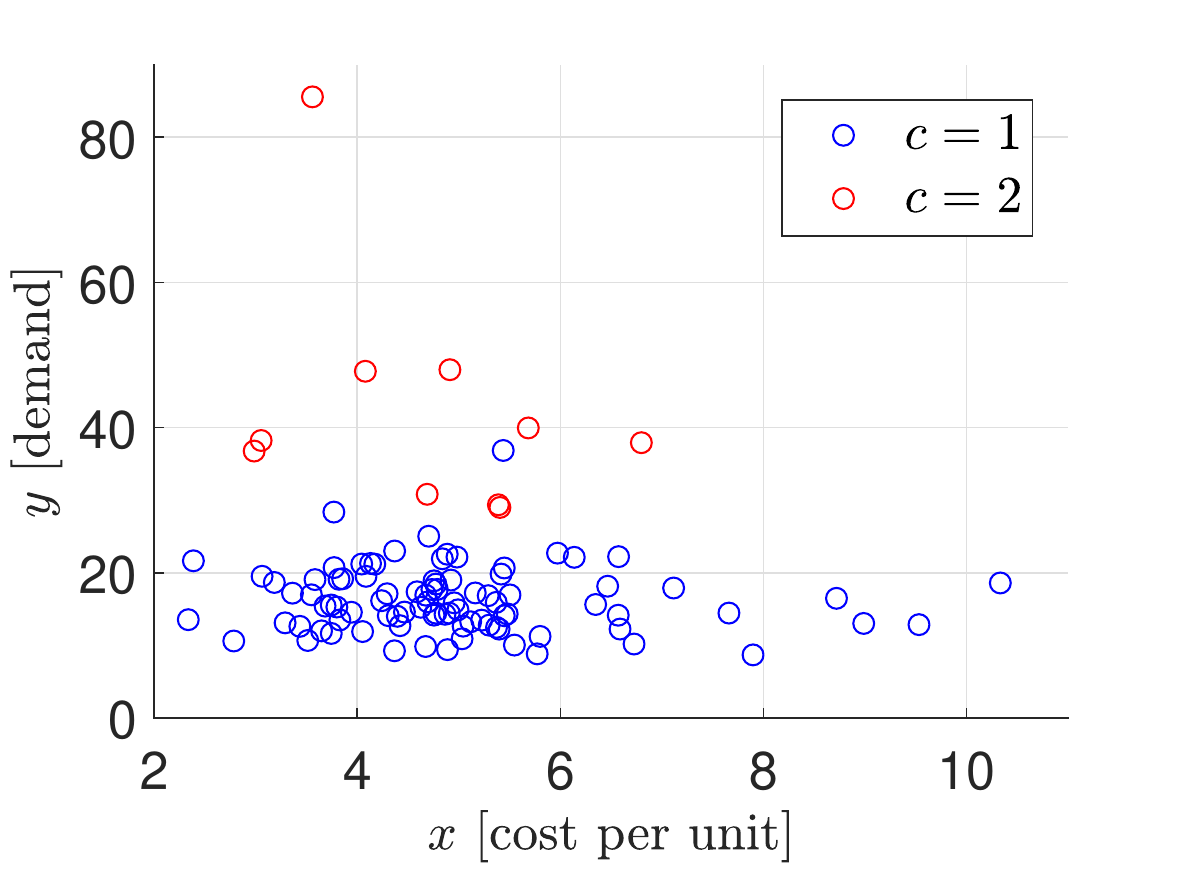}
    \caption{}
    \label{fig:supply demand data}
    \end{subfigure}
    \begin{subfigure}{0.45\linewidth}
    \includegraphics[width=0.95\linewidth]{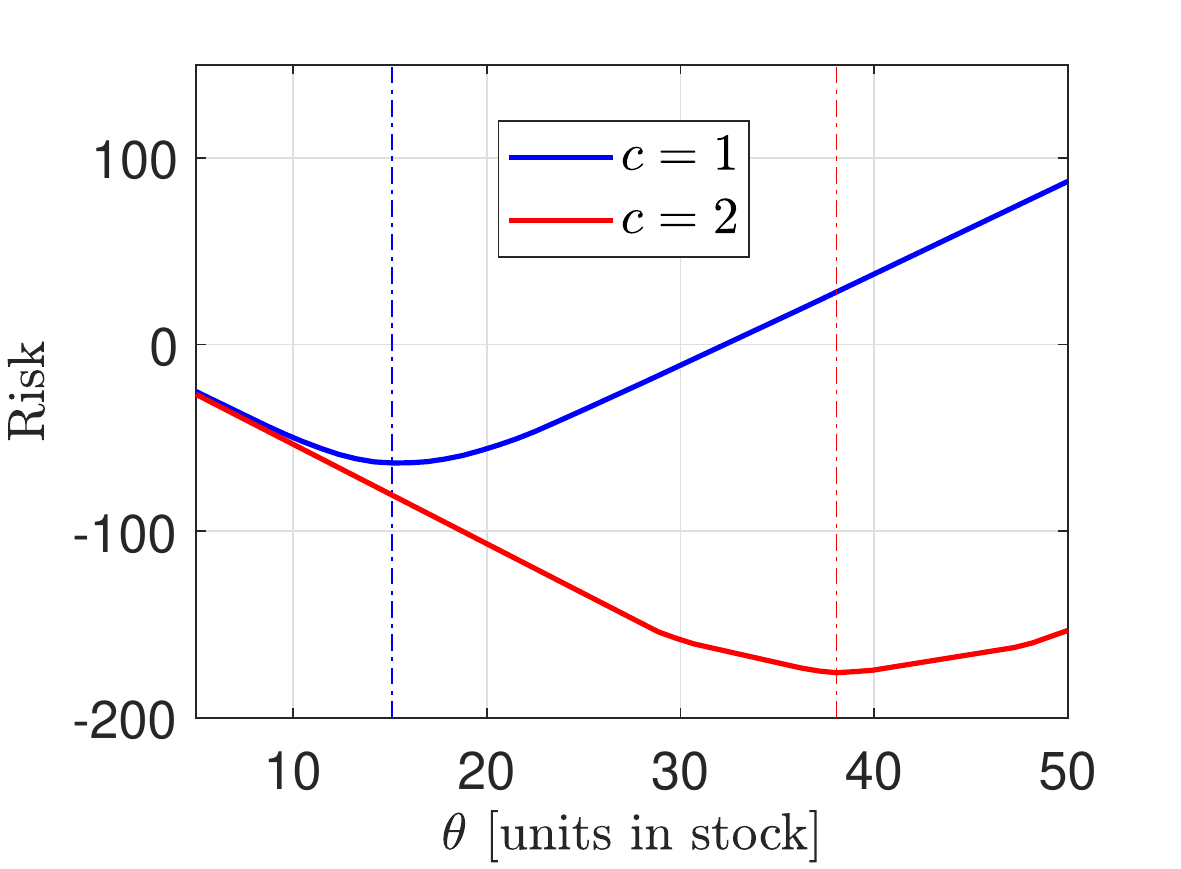}
    \caption{}
    \label{fig:supply demand loss}
    \end{subfigure}
    \caption{Stock control example. (a) Training data for cost per unit, $x$, and total demand, $y$, in two different contexts $\ctx=1$ and $\ctx=2$. The number of samples are $n_1~=~90$ and $n_2~=~10$, respectively. (b) Conditional risk $\risk_{\ctx}(\theta)$ in each context (solid lines) as a function of stock level $\theta$. Vertical lines indicate the corresponding optimal stock levels. The risk has no closed-form solution as was computed numerically using Monte Carlo simulations.}
    \label{fig: supply demand}
\end{figure*}

\begin{example}[Stock control] Suppose we must decide on the quantity $\theta$ to keep in stock for sale. Let $x$ denote the purchasing cost per unit and $y$ the total demand so that the loss of a specific stock level $\theta \in [0,\theta_{\max}]$ is
$$\lossfunc(\zvec) = \theta  x  - r \min(\theta, y) $$
where $r$ is a given price per unit and  $\zvec = (x,y)$. Figure~\ref{fig:supply demand data} illustrates past data observed in two different contexts, for instance sunny or rainy weather. Since they are heterogeneous, their conditional risks $\risk_{\ctx}(\theta)$ at stock level $\theta$ differ, as seen in Figure~\ref{fig:supply demand loss}.
\end{example}
When future contexts  are unknown, as in the example above, we  treat $\ctx$ as a random variable with a distribution $\ptrain(\ctx)$ and consider the parameter $\parm$ that minimizes the (overall) risk
\begin{equation} 
\risk(\parm; p) \equiv \sum_{\ctx \in \ctxset} \risk_{\ctx}(\parm) \prb(\ctx),
 \label{eq:totalrisk}
\end{equation}
Empirical risk minimization (\erm{}) \cite{vapnik2013nature, shapiro2014lectures} is a standard learning approach that forms an empirical approximation of \eqref{eq:totalrisk} with an \emph{unbiased} estimate of $\ptrain(\ctx)$. When the number of contexts $|\ctxset|$ is relatively large, such an unbiased estimate can readily underemphasize more challenging contexts leading to poor worst-context performance as a result.

The classical robust approach, dating back to \cite{wald1950statistical}, is the minimax  method which focuses on the worst-case risk alone. This approach was soon recognized to be overly conservative and can result in a poor trade-off between performance and robustness \cite{gabrel2014recent}.  \cite{savage1951theory} recognized that what is the `worst-case' context must be put in relation to the minimum risk achievable in that context. He therefore proposed focusing on the worst-case \emph{excess} risk (aka. regret) instead. \cite{scarf1957inventory}, on the other hand, recognized that considering a worst-case \emph{distribution} with a specified set of  distributions would avoid an overemphasis on very improbable scenarios. This approach was extended to distribution sets for $\ptrain(\zvec)$ with a size controlled by a user parameter. These sets have been constructed using a variety of measures, including moment-based constraints, Wasserstein distances, f-divergences and minimum mean discrepancy measures
\cite{delage2010distributionally,shafieezadeh2015, namkoong2017variance, blanchet2019robust,staib2019distributionally}.

In this paper, we focus on the uncertainty of $\prb(\ctx)$ using confidence-based distribution sets building upon principles congruent with those recently advanced in the decision-theoretic literature on imprecise probabilities \cite{hill2013confidence,bradley2017decision}. We combine the insights from both Savage's focus on excess risks and and Scarf's focus on distributional uncertainty to develop a method for learning from training data obtained in heterogeneous contexts. The method has the following properties:
\begin{itemize}
    \item it is robust in the worst-case contexts,
    \item computationally efficient,  
    \item uses distribution sets with a desired level of statistical confidence,
    \item and interpolates adaptively between the empirical risk minimization and minimax regret methods.  
\end{itemize}
We demonstrate the methodology in different tasks using real and synthetic data.

\section{Problem Formulation}

Using a total of $n$ samples, we seek to learn a parameter $\what{\parm}$ that approximates a minimizer of the risk $\risk(\parm; p)$, which may depend heavily on the distribution of contexts $\prb$.

\begin{example}[Stock control, cont'd] Consider two different context distributions, $\prb' = \{ 0.90, 0.10 \}$ and $\prb''=\{ 0.20, 0.80 \}$. Figure~\ref{fig: exp loss vs theta} shows how the risk $\risk(\parm; \prb)$ differs under two context distributions, along with the resulting optimal stock levels $\theta$. 
\begin{figure*}[t!]
    \centering
    \begin{subfigure}{0.45\linewidth}
    \includegraphics[width=0.9\linewidth]{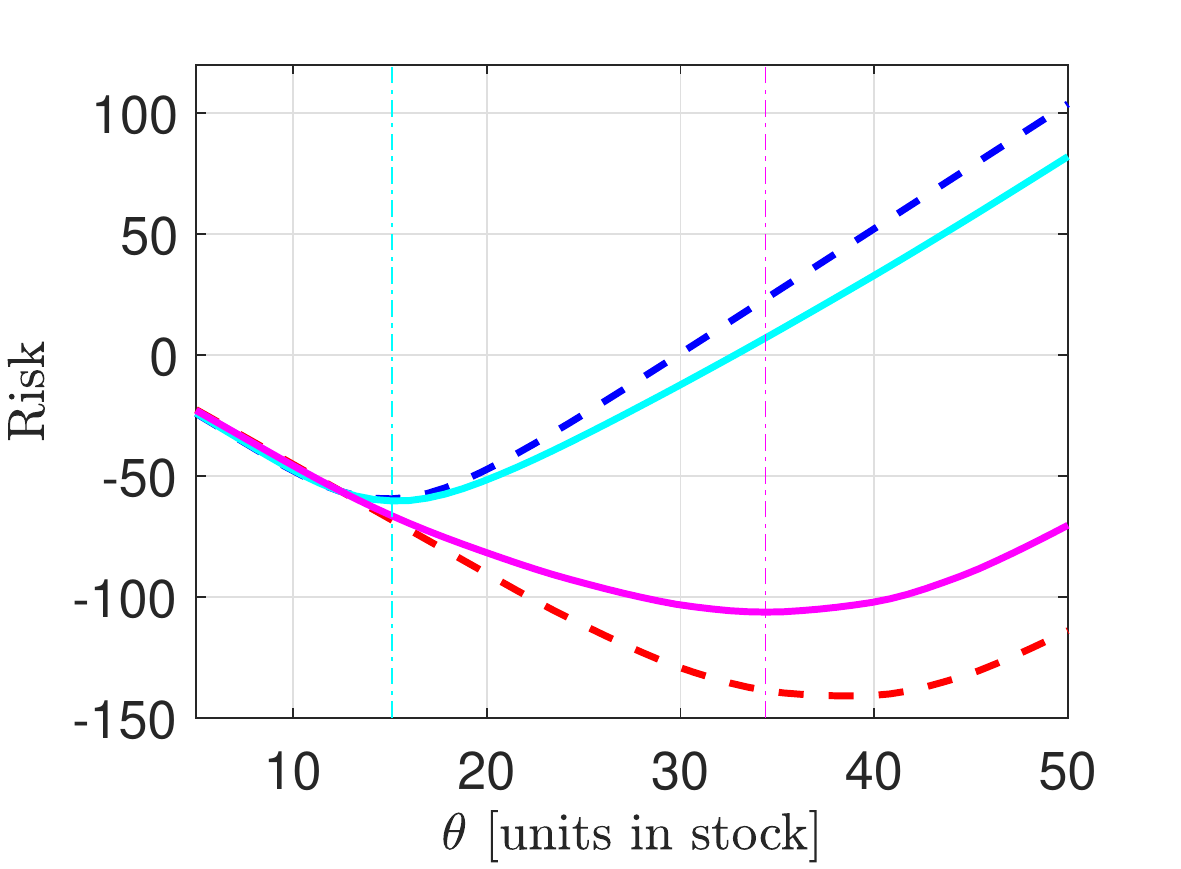}
    \caption{}
    \label{fig: exp loss vs theta}
    \end{subfigure}
    \begin{subfigure}{0.45\linewidth}
    \includegraphics[width=0.9\linewidth]{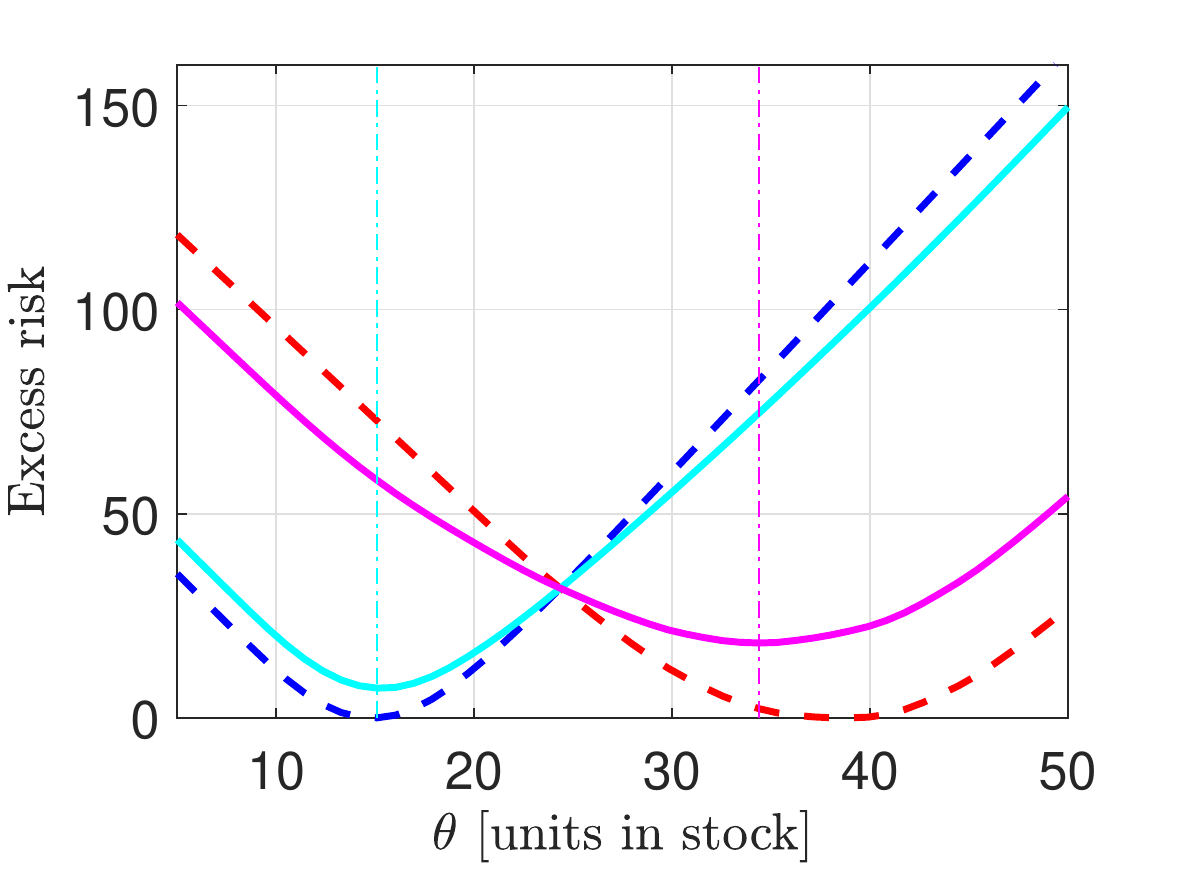}
    \caption{}
    \label{fig: regret vs theta}
    \end{subfigure}
    \caption{Stock control problem under two different context distributions, $\prb'$ and $\prb''$.
    (a) Risk $\risk(\theta; \prb)$ under $\prb'$  (solid cyan) and $\prb''$ (solid magenta). Dashed lines show the conditional risks $\risk_{\ctx}(\theta)$, see Figure~\ref{fig:supply demand loss}. (b) Excess risk $\regret(\theta; \prb)$ under $\prb'$  (solid cyan) and $\prb''$ (solid magenta). Dashed lines show the conditional excess risk in each context. Vertical lines show the optimal stock levels for $\prb'$ and $\prb''$, respectively.}
    \label{fig: prob illus part 1}
\end{figure*}
\end{example}

The empirical risk minimization approach (\erm) solves
\begin{equation} \label{eq:erm}
\min_{\parm} \; \sum_{\ctx \in \ctxset} \what{\risk}_{\ctx}(\parm) \what{\prb}(\ctx), 
\end{equation}
where $\what{\prb}(\ctx) = \frac{n_\ctx}{n}$ is an unbiased estimate of $\ptrain(\ctx)$ and $\what{\risk}_{\ctx}(\parm) = \frac{1}{n_c}\sum_i \lossfunc(\zvec_i)$ is an empirical estimate. The unbiased estimates can lead to an underemphasis on contexts with risks that are sensitive to $\parm$. Wald's robust approach focuses on the worst-case context and the (empirical) minimax risk method solves
\begin{equation} \label{eq: min max exp}
\min_{\parm} \; \max_{\ctx \in \ctxset} \what{\risk}_{\ctx}(\parm)
\end{equation}
It has two main drawbacks: it can overemphasize  highly improbable contexts and also ignore how the risk can be reduced in alternative contexts. 
For instance, in a classification task the minimax approach could focus on a context where the  error rate is high but insensitive with respect to $\parm$, while ignoring contexts with low minimum error rates that are sensitive to $\parm$. Moreover, while conceptually simple, finding a solution to \eqref{eq: min max exp}  is often a computationally challenging problem.

We now consider an alternative approach to achieve an appropriate trade-off between performance and robustness against worst-case contexts.


\section{Robust method}

The minimum achievable risk,  i.e., $\min_{\parm} \: \risk_{\ctx}(\parm)$, may differ across heterogeneous contexts. Following \cite{savage1951theory}, we focus only on the part of the risk that can be reduced by the parameter $\parm$, unlike the overly conservative nature of the approach in \eqref{eq: min max exp}. Let us now consider the  (overall) \emph{excess} risk that $\parm$ incurs, 
\begin{equation} 
 \regret(\parm; \prb) \equiv  \sum_{\ctx \in \ctxset} \: \left[ \risk_{\ctx}(\parm) - \min_{\parm'}\:  \risk_{\ctx}(\parm')  \right]  \prb(\ctx) ,
 \label{eq:expectedregret}
\end{equation}
for a given context distribution $\prb$. 

\begin{example}[Stock control, cont'd]
The excess risk $\regret(\parm;\prb)$  under two different context distributions $\prb'$ and $\prb''$ is shown in Figure~\ref{fig: regret vs theta}.
\end{example}

 The excess risk at stake for any pair $(\parm, \prb)$ is given by $\regret(\parm; \prb)$. For a given $\prb$, a parameter $\parm$ that minimizes excess risk  \eqref{eq:expectedregret} also minimizes the overall risk \eqref{eq:totalrisk}, and vice versa. In lieu of $\prb$, we consider a distribution $\prb^*$ that yield worst-case excess risk in a \emph{confidence set} of distributions, following \cite{scarf1957inventory}.

\subsection{Confidence-based learning}

Let $\prbset$ denote the set of all distributions $\prb$. Given $n$ training data points, we seek a subset $\prbset^n_{\typical} \subseteq \prbset$ that covers the unknown $\prb$ at a specified confidence level $\typical$:
\begin{equation}
\Prob\big\{  \prb \in \prbset^n_\typical \big\} \geq \typical
\label{eq:property coverage}
\end{equation}
such that
\begin{equation}
\typical < \typical' \quad \Leftrightarrow \quad  \prbset^n_{\typical} \subset \prbset^n_{\typical'} \qquad \text{and}  \qquad  \prbset^n_{1} \equiv \prbset
\label{eq:property nested}
\end{equation}
Thus the confidence level controls the size of the distribution set $\prbset^n_\typical$ at any given $n$.
\begin{result}[Confidence set] \label{result: confd set}
Let
\begin{equation}
\prbset^n_\typical = \big\{  \prb' : D(\what{\prb}\| \prb') \leq \eps_{\typical} \big\},
\label{eq:confidence set}
\end{equation}
where $D(\what{\prb}\| \prb') = \what{\E}[\log_2(\what{\prb}(\ctx)/\prb'(\ctx))]$ denotes the Kullback-Leibler divergence from $\what{\prb}(c) = n_\ctx/n$. By setting
\begin{equation}
\eps_{\typical} = \frac{1}{\ndata}\Big(|\ctxset|\log_{2}(\ndata + 1) - \log_{2}(1-\typical)\Big),
\label{eq:epstypical}
\end{equation}
the set \eqref{eq:confidence set} satisfies the coverage property \eqref{eq:property coverage} and  the nested property \eqref{eq:property nested}.
\end{result}
\begin{proof}
We have that $\{  \prb : D(\what{\prb}\| \prb) \leq \eps \} \subset \{  \prb : D(\what{\prb}\| \prb) \leq \eps' \} \Leftrightarrow \eps < \eps'$, and from \eqref{eq:epstypical} we have $\eps < \eps' \Leftrightarrow \typical < \typical'$ so that \eqref{eq:property nested} holds. Using \cite[thm.~11.2.1]{cover2012elements}, it follows that $\Prob\big\{  \prb \not \in \prbset^n_\typical \big\} \leq 2^{ |\ctxset|\log_2(n+1) -n\eps}  \equiv 1- \typical$. Thus solving for $\eps$ yields \eqref{eq:epstypical} such that \eqref{eq:property coverage} is satisfied.
\end{proof}

The tasks for which the stakes of choosing $\parm$ are high, require a confidence level $\typical$ closer to $100\%$ and thus a larger set $\prbset^n_\typical$. Conversely, for a given confidence level the evidence accumulated as $n$ increases should exclude certain context distributions. This principle has been advanced more recently in the decision-theoretic literature on imprecise probabilities, see, e.g., \cite{hill2013confidence,bradley2017decision}.

\begin{example}[Stock control, cont'd] Consider a context distribution $\prb = \{ \prb_1 , 1-\prb_1\}$. Figure~\ref{fig:illus interval pi} illustrates nested confidence sets $\prbset^n_{\typical}$ at levels 90\%, 99\% and 100\%. As the amount of training data $n$ increases, the confidence set narrows down when $\typical < 100 \%$. Here the underlying context distribution is $\ptrain=\{ 0.95, 0.05 \}$.

\begin{figure*}[t!]
    \centering
        \begin{subfigure}{0.45\linewidth}
    \includegraphics[width=0.95\linewidth]{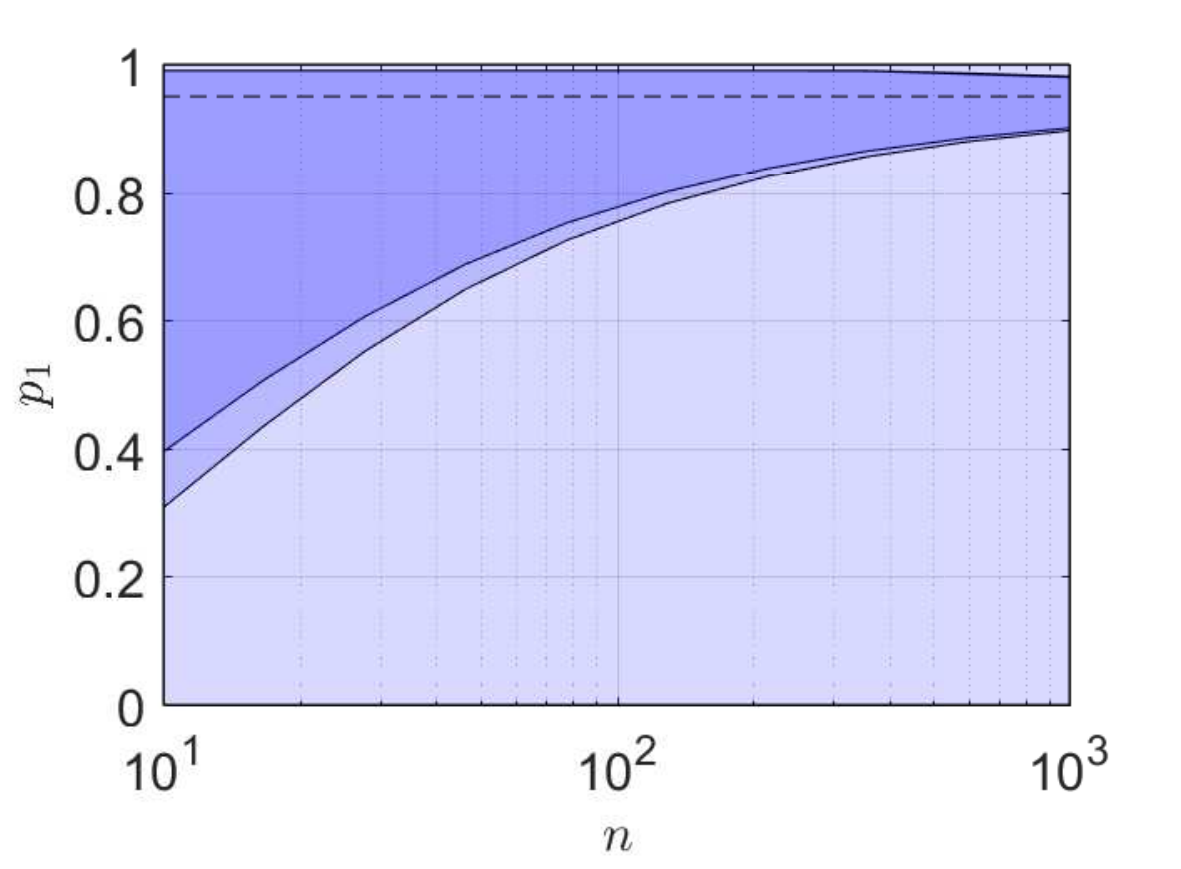}
    \caption{}
    \label{fig:illus interval pi}
    \end{subfigure}
    \begin{subfigure}{0.45\linewidth}
    \includegraphics[width=0.95\linewidth]{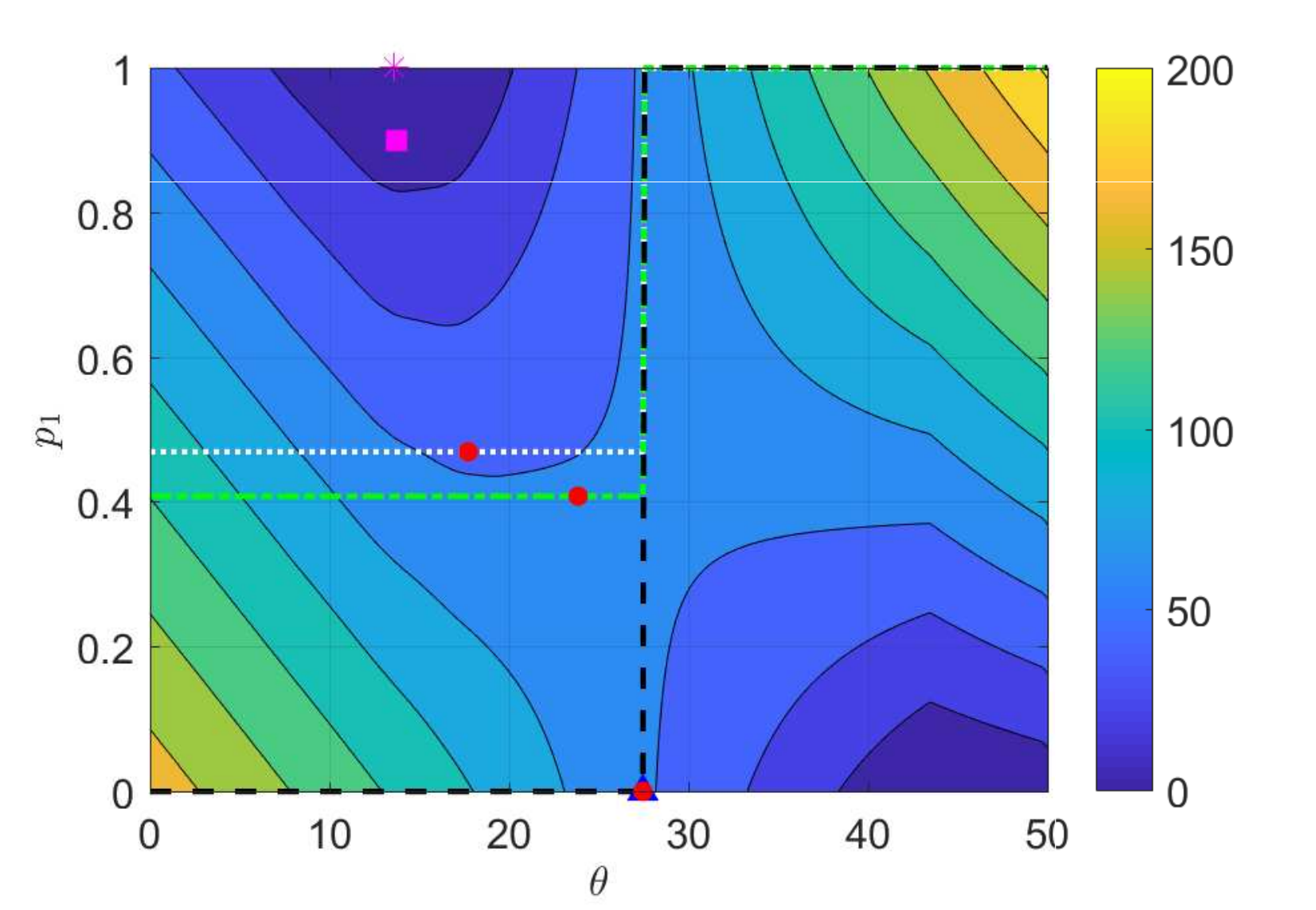}
    \caption{}
    \label{fig:illus contours}
    \end{subfigure}
    \caption{Stock control problem with increasing amounts of training data $n$ and unknown  context distribution $\prb = \{ \prb_1, 1-\prb_1\}$. (a) Confidence sets $\prbset_{\typical}^{\ndata}$ with levels $90\%$, $99\%$ and $100\%$ (shaded), which are nested according to \eqref{eq:property nested}. Note that at $\typical = 100\%$, the set contains all distributions. Dashed line shows the unknown probability $\prb_1$. (b) Contour plot of empirical excess risk $\regrethat(\theta; \prb)$ when $\ndata~=~20$. The lines show the least favourable $\prb_{1}$ for each $\theta$, obtained from the inner problem in \eqref{eq: main opt prob}, using three confidence levels $\typical$:  90\% (dotted white), 99\% (dashed cyan) and 100\% (solid black). The red dots on the lines denote the corresponding minimizers $\what{\theta}_{\typical}$ of \eqref{eq: main opt prob}. For comparison, we include \erm{}
\eqref{eq:erm} (square) and minimax risk $\theta$ \eqref{eq: min max exp} (star).}
    \label{fig: prob illus part 2}
\end{figure*}

\end{example}

We now consider $\parm$ that minimizes worst-case excess risk over all distributions in the confidence set $\prbset^n_\typical$ to overcome the two main drawbacks of the minimax risk approach \eqref{eq: min max exp}. That is,
\begin{equation} \label{eq: main opt prob}
\boxed{\parmdrr \in \argmin_{\parm} \: \max_{\prb \in \prbset^n_\typical} \; \expregrethat(\parm; \prb)}
\end{equation}
using estimates the conditional risks in \eqref{eq:expectedregret}. By increasing the level $\typical$, we are increasingly confident that we cover the unknown $\prb$ in \eqref{eq:property coverage}. 

\begin{example}[Stock control, cont'd] Figure~\ref{fig:illus contours} shows the empirical excess risk $\expregrethat(\theta; \prb)$ for different stock levels $\theta$ and context distributions $q = \{ \prb_1, 1-\prb_1\}$. \erm{} and the minimax risk methods give overwhelming weight to context $\ctx=1$. By contrast, as the confidence level $\beta$ increases, the robust approach takes  the excess risk of $\theta$ in context $\ctx=2$ increasingly into account so that \eqref{eq: main opt prob} approaches $\theta \approx 25$ at which the excess risks for each context are equal. Note that for $\theta$ lower (or greater) than this level, the least-favorable distribution $\prb^*$ gives maximum weight to $\ctx=2$ (or, alternatively $\ctx=1$).
\end{example}



We now show that the approach in \eqref{eq: main opt prob} interpolates adaptively between empirical risk minimization and minimax regret depending on the confidence level $\typical$.

\begin{result} \label{result: opt weights}
Let $\prb^{*}$ be the least-favorable distribution in \eqref{eq: main opt prob}  and
$$\regrethat_{\ctx}(\parm) = \what{\risk}_{\ctx}(\parm) - \min_{\parm'}~\what{\risk}_{\ctx}(\parm') \geq 0$$
denote the empirical excess risk for context $\ctx$. Then the cost function in \eqref{eq: main opt prob} is equivalent to a weighted combination of empirical risk and excess risk:
\begin{equation} \label{eq: regret and erm}
\boxed{ \regrethat(\parm; \prb^{*}) = \what{\risk}(\parm; \what{\prb}) + \sum_{\ctx}\what{\prb}(\ctx)\wt_{\ctx}^{\typical}(\parm) \regrethat_{\ctx}(\parm) + K,   } 
\end{equation}
where $K$ is a constant and the weights $\wt_{\ctx}^{\typical}(\parm) \in [-1,1]$. 
Specifically,
\begin{equation} \label{eq: w wts}
    \wt_{\ctx}^{\typical}(\parm)= \Bigg(\sum_{\ctx'}\what{\prb}(\ctx')\frac{\nu^{*}-\regrethat_{\ctx}(\parm)}{\nu^{*}-\regrethat_{\ctx'}(\parm)}\Bigg)^{-1} - 1
\end{equation}
where $\nu^* > \max_{\ctx}\regrethat_{\ctx}(\parm)$ is the root of the following equation
\begin{equation} \label{eq:nu}
\begin{split}
\sum_{\ctx\in\ctxset} \what{\prb}(\ctx)\log_2\left(\nu^{*} - \regrethat_{\ctx}(\parm)\right)+&
\log_2\left[\sum_{\ctx\in\ctxset}\frac{\what{\prb}(\ctx)}{\nu^{*} - \regrethat_{\ctx}(\parm)}\right] \\- &\eps_{\typical} = 0.
\end{split}
\end{equation}
Furthermore, we have that
\begin{enumerate}
    \item as $n\rightarrow\infty$ for a given $\typical$:  $$\regrethat(\parm; \prb^{*})\rightarrow\what{\risk}(\parm; \what{\prb}) + K$$ 
    \item as $\typical\rightarrow 1$, for a given $n$:
    $$\regrethat(\parm;\prb^{*})\rightarrow \max_{\ctx} \regrethat_{\ctx}(\parm)$$
\end{enumerate}
That is, in the limiting cases \eqref{eq: main opt prob} approaches \erm{} as $n$ become large and it approaches an empirical minimax regret method with increasing confidence level $\typical$. 
\end{result}

\begin{proof}
The Lagrangian of the inner maximization problem of \eqref{eq: main opt prob} is, 
\begin{equation}
\begin{split}
    L(\prb, \mbs{\lambda}, \nu) &= -\sum_{\ctx \in \ctxset}\regrethat_{\ctx}(\parm)\prb_{\ctx} + \lambda_{0} [D(\what{\prb}||\prb) - \eps_{\typical}] +\\ 
    &\sum_{\ctx \in \ctxset}\lambda_{\ctx}(-\prb_{\ctx})  + \nu\left(\sum_{\ctx \in \ctxset} \prb_{\ctx} - 1\right).
\end{split}
\end{equation}
We have $\what{\prb}_c = n_c/n > 0$ for all $c$. When $\prb \equiv \what{\prb}$, the inequality constraints are inactive and Slater's condition is satisfied, which implies that the primal-dual optimal solutions $(\prb^{*}, \mbs{\lambda}^{*}, \nu^{*})$ can be obtained by solving the following Karush-Kuhn-Tucker (KKT) conditions:
\begin{equation}
\begin{split}
    \partial_{\prb_{\ctx}} L({\prb}^{*}, \mbs{\lambda}^{*}, \nu^{*})&=0, \quad \lambda_{\ctx}^{*}\prb_{\ctx}^{*}=0,\quad \prb_{\ctx}^{*}\geq 0~~\text{for}~~\ctx~\in~\ctxset,\\
    \lambda_{0}^{*} \left[D(\what{\prb}||\prb^{*}) - \eps_{\typical}\right]&=0,\quad \sum_{\ctx} \prb^*_c =1,\quad D(\what{\prb}||\prb^{*})\leq \eps_{\typical}
\end{split}
\end{equation}
From the derivative condition on the Lagrangian, we obtain
\begin{equation}
    \prb_{\ctx}^{*}=\frac{\lambda_{0}^{*}\what{\prb}_\ctx}{\nu^{*} -  \regrethat_{\ctx}(\parm)}~~\text{for}~~\ctx\in\ctxset.
\label{eq:opt prb weights}
\end{equation}
Since $\prb_{\ctx}^{*}>0$ it follows that $\lambda_{\ctx}^{*}~=~0$. Moreover, $\lambda_{0}^{*} > 0$ since it is a common scale factor and $\lambda_{0}^{*} = (\sum_{\ctx} \frac{\what{\prb}_{\ctx}}{\nu^{*} - \regrethat_{\ctx}(\parm)})^{-1}$ is obtained from $\sum_\ctx \prb^*_\ctx = 1$. It follows that $D(\what{\prb}||\prb^{*})-  \eps_{\typical} = 0$, for which $\nu^{*}$ is a root. 

Now let $\what{\risk}(\parm)$ be the empirical risk in \eqref{eq:erm},  $\what{r}_{\ctx}~=~\min_{\parm} \what{\risk}_{\ctx}(\parm)$ be the minimum risk in context $\ctx$ and $\regrethat_{\ctx}(\parm)~=~\what{\risk}_{\ctx}(\parm) - \what{r}_{\ctx}$, then  
\begin{equation}
    \regrethat(\parm;\prb^{*}) - \what{\risk}(\parm) = \sum_{\ctx}\prb^{*}(\ctx) \regrethat_{\ctx}(\parm) - \sum_{\ctx}\what{\prb}(\ctx)\what{\risk}_{\ctx}(\parm).
\end{equation}
Plugging $\prb^{*}(\ctx)$ from \eqref{eq:opt prb weights} gives
\begin{equation}
     \regrethat(\parm;\prb^{*}) - \what{\risk}(\parm) = \sum_{\ctx}\what{\prb}(\ctx)\Bigg( \regrethat_{\ctx}(\parm)\frac{\lambda_{0}^{*}}{\nu^{*} -  \regrethat_{\ctx}(\parm)} -\what{\risk}_{\ctx}(\parm)\Bigg).
\end{equation}
Adding and subtracting $\sum_{\ctx}\what{\prb}(\ctx)\what{r}_{\ctx}$ on the right hand side we get,
\begin{equation}
     \regrethat(\parm;\prb^{*}) - \what{\risk}(\parm) = \sum_{\ctx}\what{\prb}(\ctx) \regrethat_{\ctx}(\parm)\Bigg(\frac{\lambda_{0}^{*}}{\nu^{*} -  \regrethat_{\ctx}(\parm)} -1\Bigg) + K.
\end{equation}
Let 
\begin{equation}
    \wt_{\ctx}^{\typical}(\parm) = \frac{\lambda_{0}^{*}}{\nu^{*} -  \regrethat_{\ctx}(\parm)} -1.
\end{equation}
Plugging in $\lambda_{0}^{*}$ in the above equation and simplifying gives \eqref{eq: w wts}.

In the first case,  $\eps_\typical \rightarrow 0$ as $n\rightarrow \infty$ using \eqref{eq:epstypical}. Then $\nu^{*}\rightarrow\infty$  \eqref{eq:nu} and from \eqref{eq: w wts}, it can be seen that the fractions in the denominator cancel in the limit and $\wt_{\ctx}^{\typical}(\parm)\rightarrow(\what{\prb}(\ctx) + \sum_{\ctx',~\ctx'\neq\ctx}\what{\prb}(\ctx'))^{-1} - 1~=~0~\forall~\ctx$. Hence, $\regrethat(\parm;\prb^{*})\rightarrow\what{\risk}(\theta) + K$ in \eqref{eq: regret and erm}. In the second case when $\typical\rightarrow 1$, $\eps_{\typical}\rightarrow\infty$ and $\nu^{*}$ approaches $\max_{\ctx}\regrethat_{\ctx}(\parm)$. Then from \eqref{eq: w wts}, $\wt_{\ctx}^{\typical}(\parm)\rightarrow~-1~\forall~\ctx$ except the context with maximum regret for which it approaches $\frac{1 - \what{\prb}(\ctx)}{\what{\prb}(\ctx)}$. Plugging these in \eqref{eq: regret and erm} gives $\regrethat(\parm;\prb^{*})\rightarrow\max_{\ctx}\regrethat_{\ctx}(\parm)$.


\end{proof}

\subsection{Numerical search method}
Learning problems typically require using numerical search methods. Here we propose using gradient-based approaches to solve  the problem \eqref{eq: main opt prob}. Let
\begin{equation}
\f(\parm)=\max_{\prb\in\prbset_{\typical}^{n}} \regrethat(\parm; \prb)
\label{eq:wcexcessrisk}
\end{equation}
In case of loss functions for which the expected losses are convex and differentiable in $\parm$, e.g., squared-error loss or cross-entropy loss for models that are linear in the parameters, we follow the approach in \cite{hu2018does}. Then the gradient of \eqref{eq:wcexcessrisk} can be obtained using Danskin's theorem \cite{danskin1966theory}: $\partial_{\parm} \f(\parm)|_{\parm'} \equiv \partial_{\parm} \regrethat(\parm; \prb^*)|_{\parm'}$,
where $\prb^*$ maximizes $\regrethat(\parm', \prb)$ and be computed efficiently by \eqref{eq:opt prb weights} evaluated at $\parm'$. We can therefore readily use gradient-based algorithms to compute $\parmdrr$, such as Algorithm~\ref{algo: gd danskin}.
\begin{algorithm}
\caption{Gradient descent algorithm} \label{algo: gd danskin}
\begin{algorithmic}[1]
\STATE Input: data $\{(\zvec_i, \ctx_{i})\}_{i=1}^{\ndata}$, confidence level: $\typical$, initial estimate: $\parm' = \parmerm$, step size : $\eta$
\STATE \textbf{repeat}
\STATE Find solution $\nu^*$ of \eqref{eq:nu}
\STATE Compute $\prb^*$ given by \eqref{eq:opt prb weights}
\STATE Compute gradient $\partial_{\parm} \f(\parm)\Big|_{\parm'}~=~\sum_{\ctx}\frac{\prb^*_{\ctx}}{\ndata_\ctx} \sum^{\ndata_\ctx}_{i=1}[\partial_{\parm}\lossfunc(\zvec_i)]\Big|_{\parm'}$
\STATE Update $\parm'~=~\parm' - \eta\partial_{\parm} \f(\parm)\Big|_{\parm'}$ 
\STATE \textbf{until convergence}
\STATE Output: $\parmdrr~=~\parm'$
\end{algorithmic}
\end{algorithm}

For cases where the loss function $\lossfunc(\zvec)$ is not convex in $\parm$, an alternative method is to use the two-time scale gradient descent-ascent algorithm, which converges to a local Nash equilibrium \cite{heusel17}. 
 


\section{Experiments} \label{sec: experminents}

We illustrate the proposed robust method for two different tasks -- stock control and classification -- and evaluate the excess risk it incurs across contexts. The data is drawn as \eqref{eq:data} with a context distribution $\prb(\ctx)$. We consider two extremes: (i) \emph{nominal case} such that $\prb$ will be covered by $\prbset^n_{\typical}$ with probability $\typical$ and (ii) \emph{worst case} in which $\prb$ is the least-favourable distribution in the entire set $\prbset$. This corresponds to focusing on the context that yields the worst excess risk. Throughout all experiments we use a confidence level of $\typical=99\%$.

\begin{figure*}[t!]
    \centering
    \begin{subfigure}{0.45\linewidth}
    \includegraphics[width=0.95\linewidth]{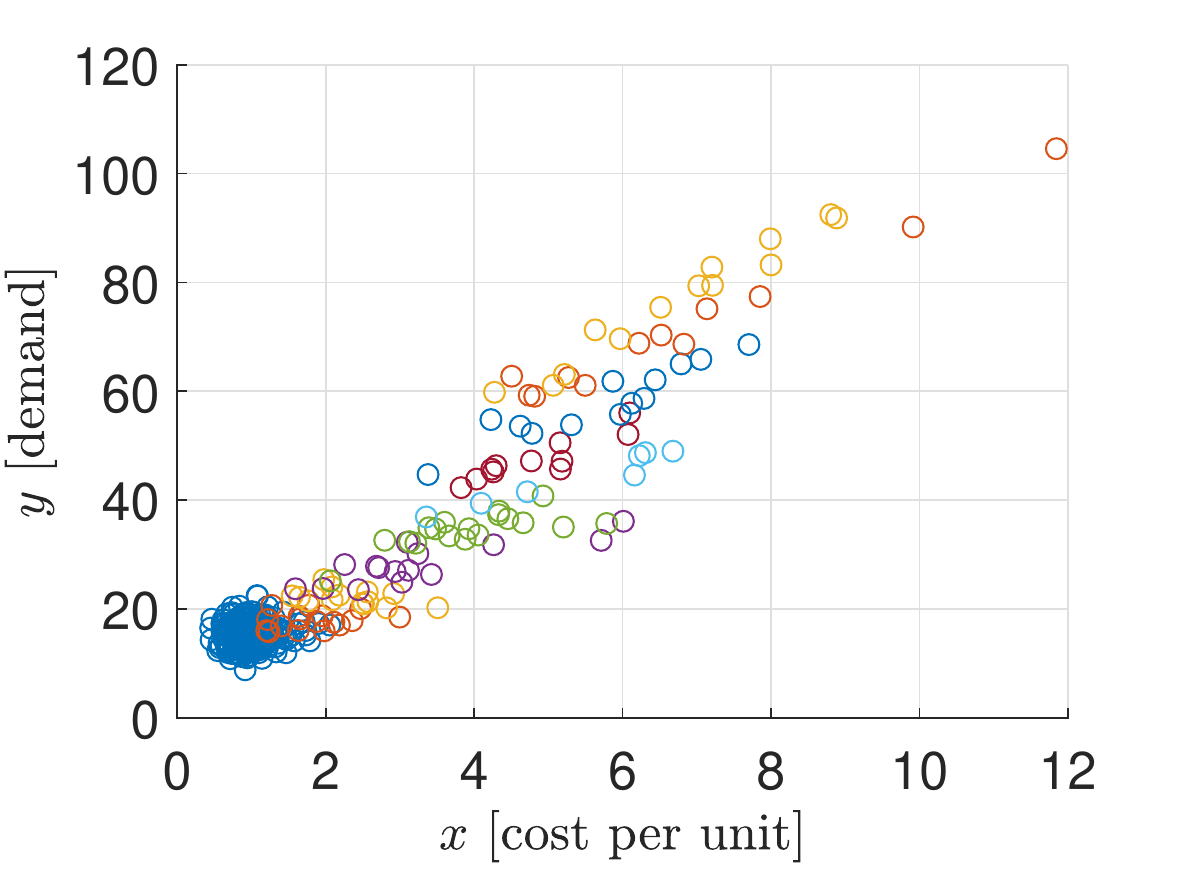}
    \caption{}
    \label{fig:ex supply demand data}
    \end{subfigure}
    \begin{subfigure}{0.45\linewidth}
    \includegraphics[width=0.95\linewidth]{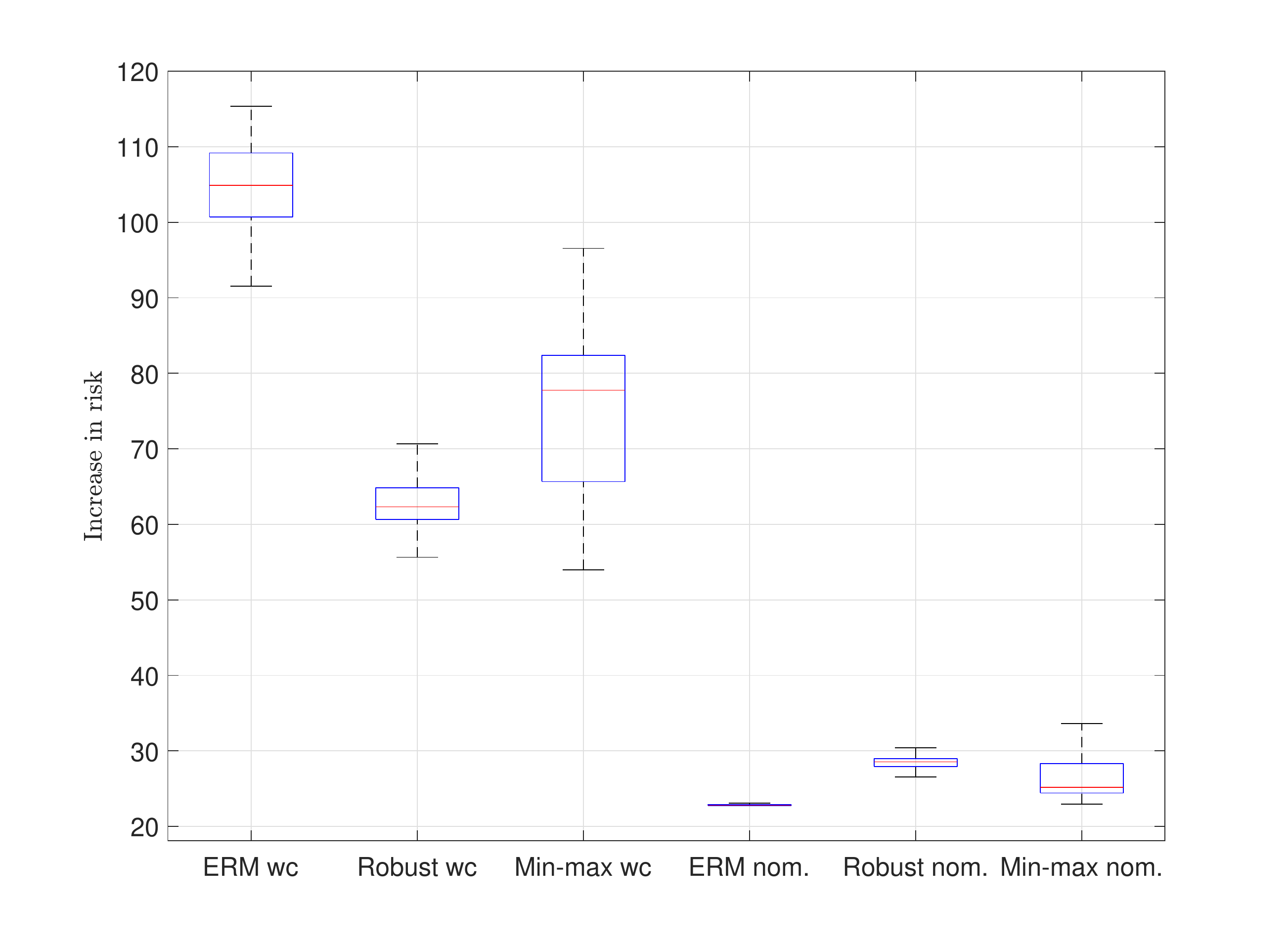}
    \caption{}
    \label{fig:ex supply demand risk}
    \end{subfigure}
    \caption{Stock control task: (a) Training data obtained from $10$ different contexts (indicated using different colours). Data for context $1$ is most prevalent while data for other contexts are scarce. (b) Box plot of excess risk for \erm~, minimax  and robust methods under worst-case (wc) and nominal (nom) test scenarios, using training data from $50$ Monte Carlo runs.}
    \label{fig:ex supply demand}
\end{figure*}

\begin{figure*}[t!]
    \centering
    \begin{subfigure}{0.45\linewidth}
    \includegraphics[width=0.9\linewidth]{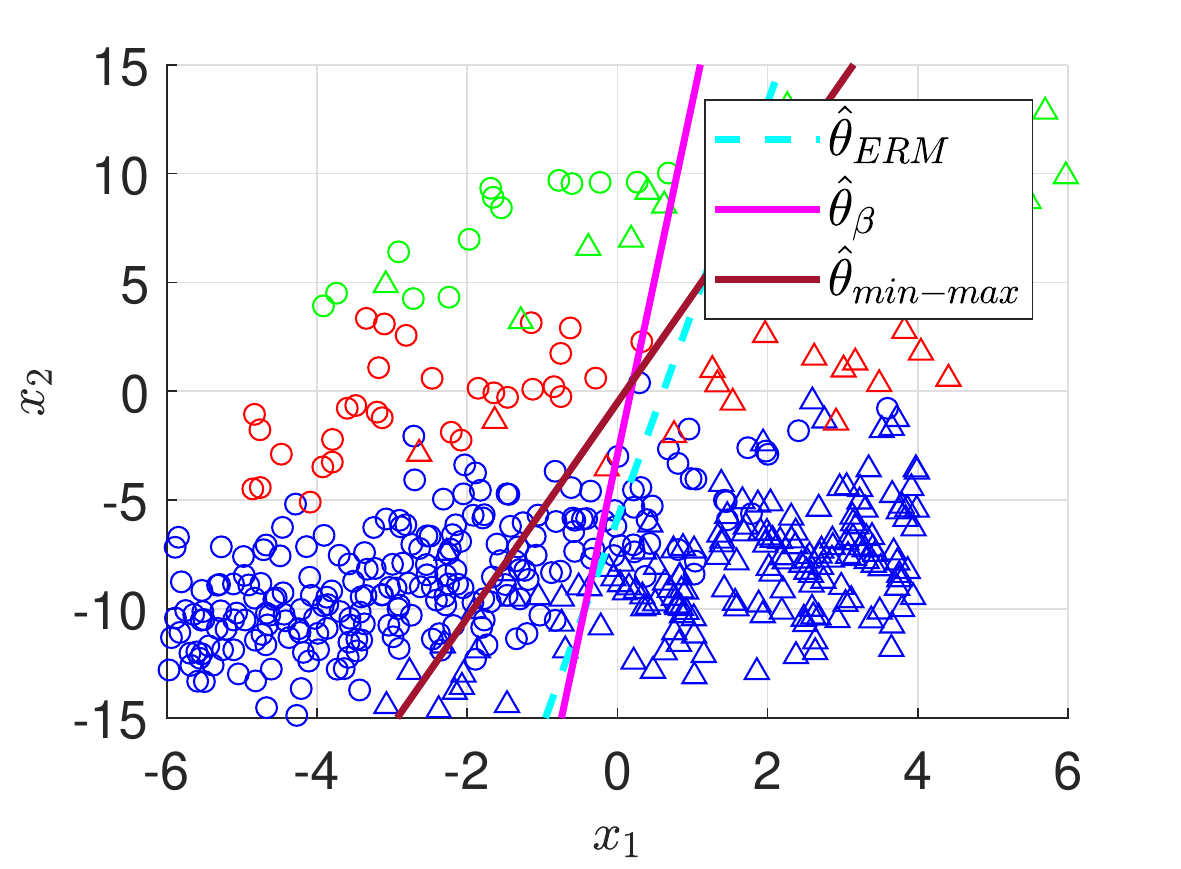}
    \caption{}
    \label{fig:class theta_hat}
    \end{subfigure}
    \begin{subfigure}{0.45\linewidth}
    \includegraphics[width=0.9\linewidth]{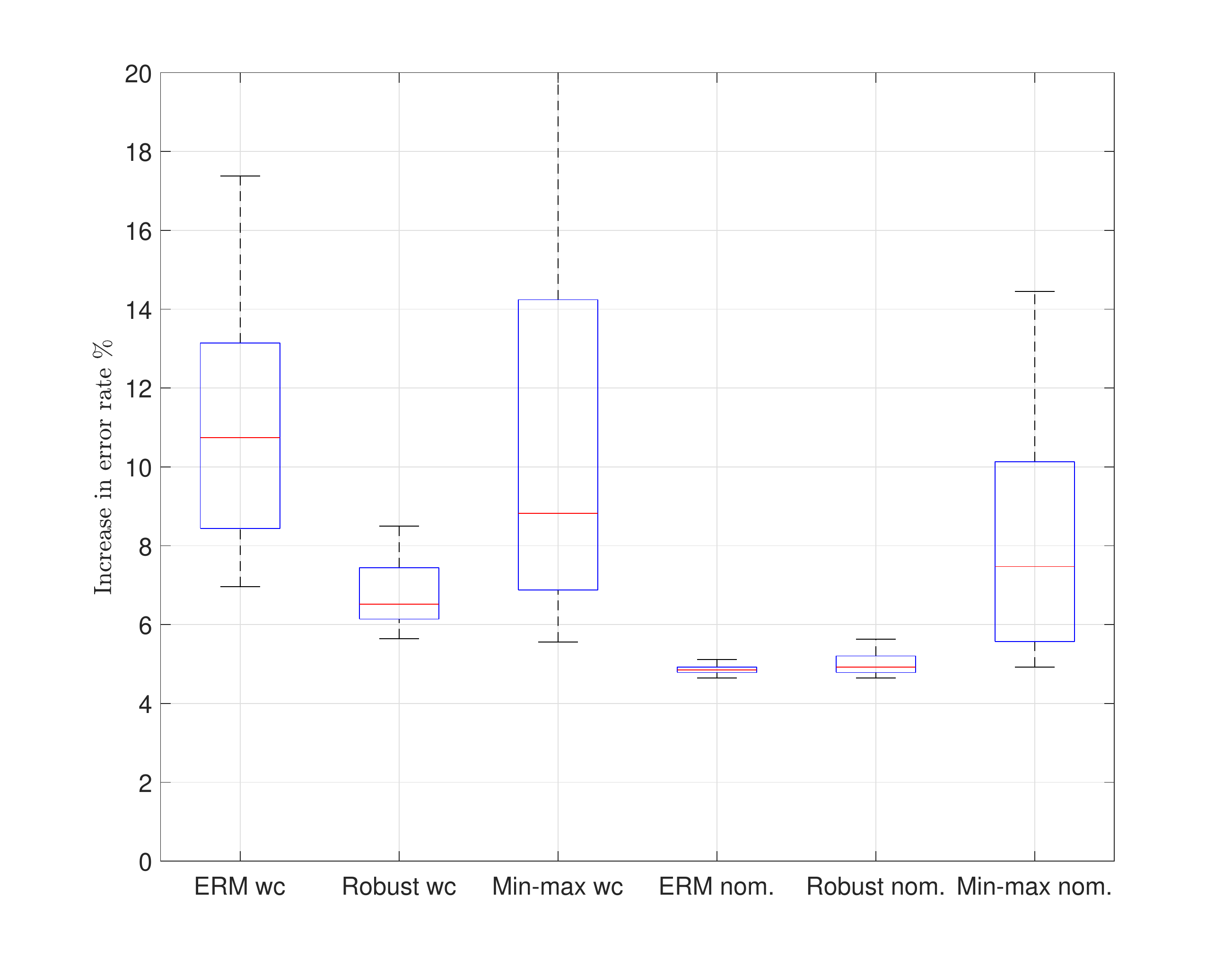}
    \caption{}
    \label{fig:class test acc}
    \end{subfigure}
    \caption{Classification task: (a) Training data from 3 different contexts (different colours) with label $\y~=~0$ (circle) and $\y~=~1$ (triangle). Plot also shows learned decision-boundaries parameterized by $\parm$. (b) Box plot of excess error rates under worst-case (wc) and nominal (nom) test conditions, using training data from $50$ Monte Carlo runs.}
    \label{fig: Classification}
\end{figure*}


\subsection{Stock control task}
We consider the stock control problem again but with $|\ctxset| = 10$ different contexts. We obtain $\ndata~=~400$ samples on the price per unit $\x$, toal demand $\y$ and context $\ctx$, drawn as
\begin{equation}
\begin{split}
    \ctx&\sim\text{Cat}(p_1,\dots, p_{|\ctxset|}),\quad
    \x|\ctx\sim\text{Log-Normal}(\mu_\ctx, 0.25),\\
    \y|\x, \ctx &\sim~ \text{Normal}(a_\ctx\x + b_\ctx, 4),
\end{split}
\end{equation}
where the probability of the first context is $p_1 = 0.70$ and the remaining contexts have equal probabilities. The training data is illustrated in Figure~\ref{fig:ex supply demand data}, using coefficients
\begin{equation}
\begin{split}
\mu_\ctx &= \frac{6}{|\ctxset|-1}(\ctx-1) + 1,\quad
    a_\ctx=\frac{6.9}{|\ctxset|-1}(\ctx-1) + 0.1,\\
    b_\ctx &=\frac{15}{|\ctxset|-1} (\ctx-1) + 15
\end{split}
\end{equation}

We compare the proposed method with the \erm~ and minimax risk methods \eqref{eq:erm}  and \eqref{eq: min max exp}. The increased risk that $\what{\theta}_{\text{erm}}$, $\what{\theta}_\text{min-max}$ and $\what{\theta}_{\typical}$ incur over all contexts is shown in Figure \ref{fig:ex supply demand risk} using $50$ different draws of training data. We see that in the worst case, the robust method has an excess risk around 60, as compared to 105 for \erm{} and 80 for minimax. This robustness is gained at the minor expense of raising the excess risk in the nominal case to about 28 as compared to 23 for \erm{} and 25 for minimax.

\subsection{Classification task}
We consider a binary classification problem with two dimensional covariates $\xvec$ observable in $|\ctxset|=3$ contexts. For the training data, the context probabilities are $\ptrain~=~\{0.8,~0.1,~0.1\}$ so that context $\ctx=1$ is most prevalent.

We obtain $\ndata~=~1000$ samples, drawn as 
\begin{equation} \label{eq: class datagen}
\begin{split}
    &\ctx\sim  ~\text{Cat}(\ptrain_1, \dots, \ptrain_{|\ctxset|}),\quad \x_{1}|\ctx \sim ~\text{Unif}\big([-5 + \mu_\ctx,~5 + \mu_\ctx]\big),\\ &\y|x_1,\ctx\sim~\text{Binomial}(\sigma(\x_{1})),\\
    &\x_{2}|x_1,y, \ctx \sim \mathcal{N}( \x_{1} + a_\ctx + 2~(\mbf{1}\{\y=0\} - \mbf{1}\{\y=1\}), 4 ),
\end{split}
\end{equation}
where $\sigma(\cdot)$ is the logistic function. The training data is illustrated in Figure \ref{fig:class theta_hat} when the coefficients are $\mu_{\ctx} \in \{  -1,0,1 \}$   and $a_\ctx \in \{ -8,0,8 \}$ corresponding to $\ctx=1,2,3$. Note that given $x_1$, the covariate $x_2$ provides no additional information about the outcome $y$. For the classification task, we use logistic regression model with cross-entropy loss function, i.e.,
\begin{equation*}
        \lossfunc(\xvec, \y)=-\y\ln\left(\sigma(\xvec^\top\parm)\right) -
        (1-\y)\ln\left(1-\sigma(\xvec^\top \parm)\right),
\end{equation*}
The minimum error rates for the three contexts were $6.86\%,~7.42\%$ and $6.70\%$ respectively.
 
We compare our method with \erm{} and minimax risk methods in \eqref{eq:erm} and \eqref{eq: min max exp} respectively. For the minimax method we use the algorithm proposed in \cite{Sagawa*2020Distributionally} which updates $\prb$ and $\parm$ iteratively and also has a convergence guarantee. No explicit guideline is provided in \cite{Sagawa*2020Distributionally} for choosing the step sizes and number of iteration parameters. We used a large number of iterations $T~=~2 \times  10^4$ iterations and step sizes of $0.1$. Figure \ref{fig:class theta_hat} shows the learned decision boundaries parameterized by $\parm$. We see that $\parmdrr$ accounts for excess risks in less frequent contexts by a more vertical decision boundary than $\parmerm$ and $\parmmm$. Since the cross entropy risk is taken as a proxy for test error, we evaluate the learning methods with respect to excess test error rate across contexts in Figure~\ref{fig:class test acc}, using 50 Monte Carlo runs. In comparison with \erm{} and minimax risk parameters, the robust method significantly reduces the excess error rate in the worst case, with a minor increase in the nominal case. In Algorithm \ref{algo: gd danskin}, we used a step size $\eta~=~0.05$ to obtain $\parmdrr$.

\begin{figure*}[t!]
    \centering
    \begin{subfigure}{0.45\linewidth}
    \includegraphics[width=0.9\linewidth]{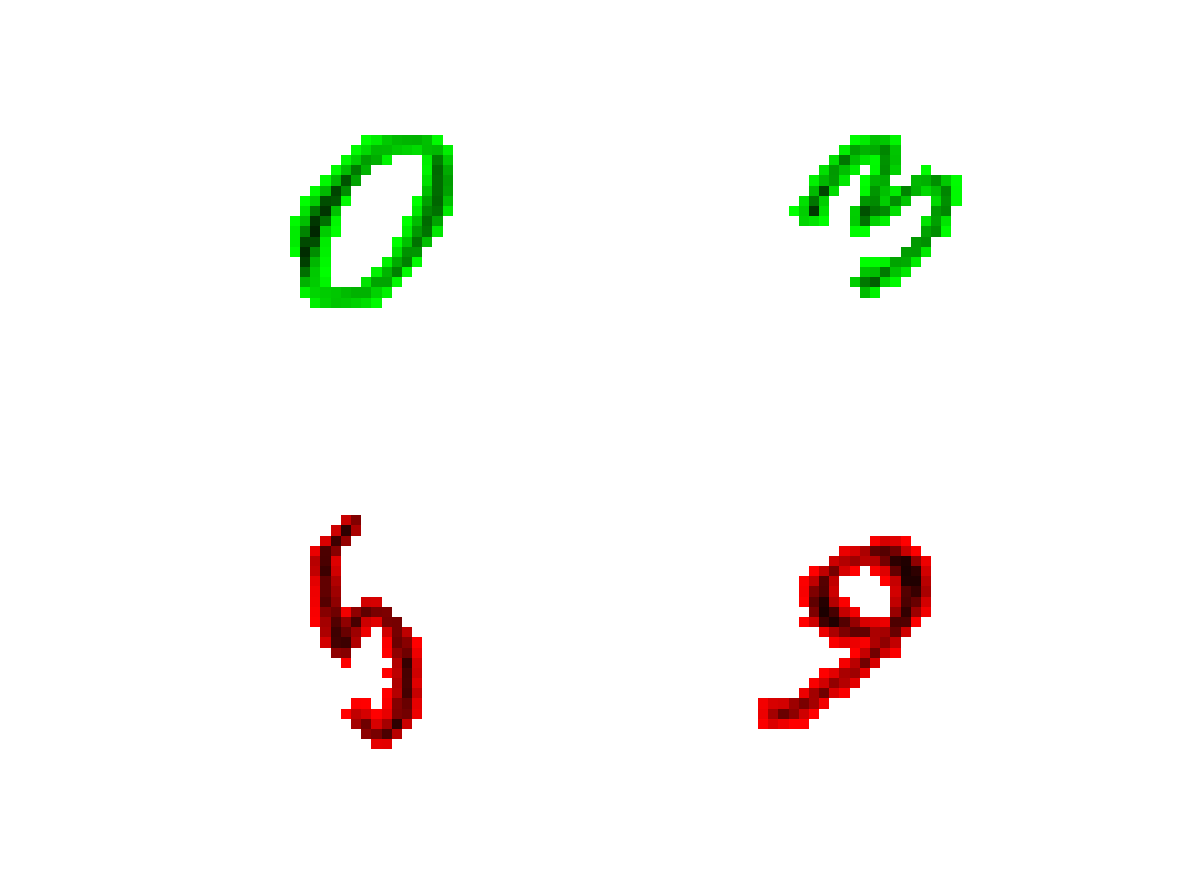}
    \caption{}
    \label{fig:mnist digits}
    \end{subfigure}
    \begin{subfigure}{0.45\linewidth}
    \includegraphics[width=0.9\linewidth]{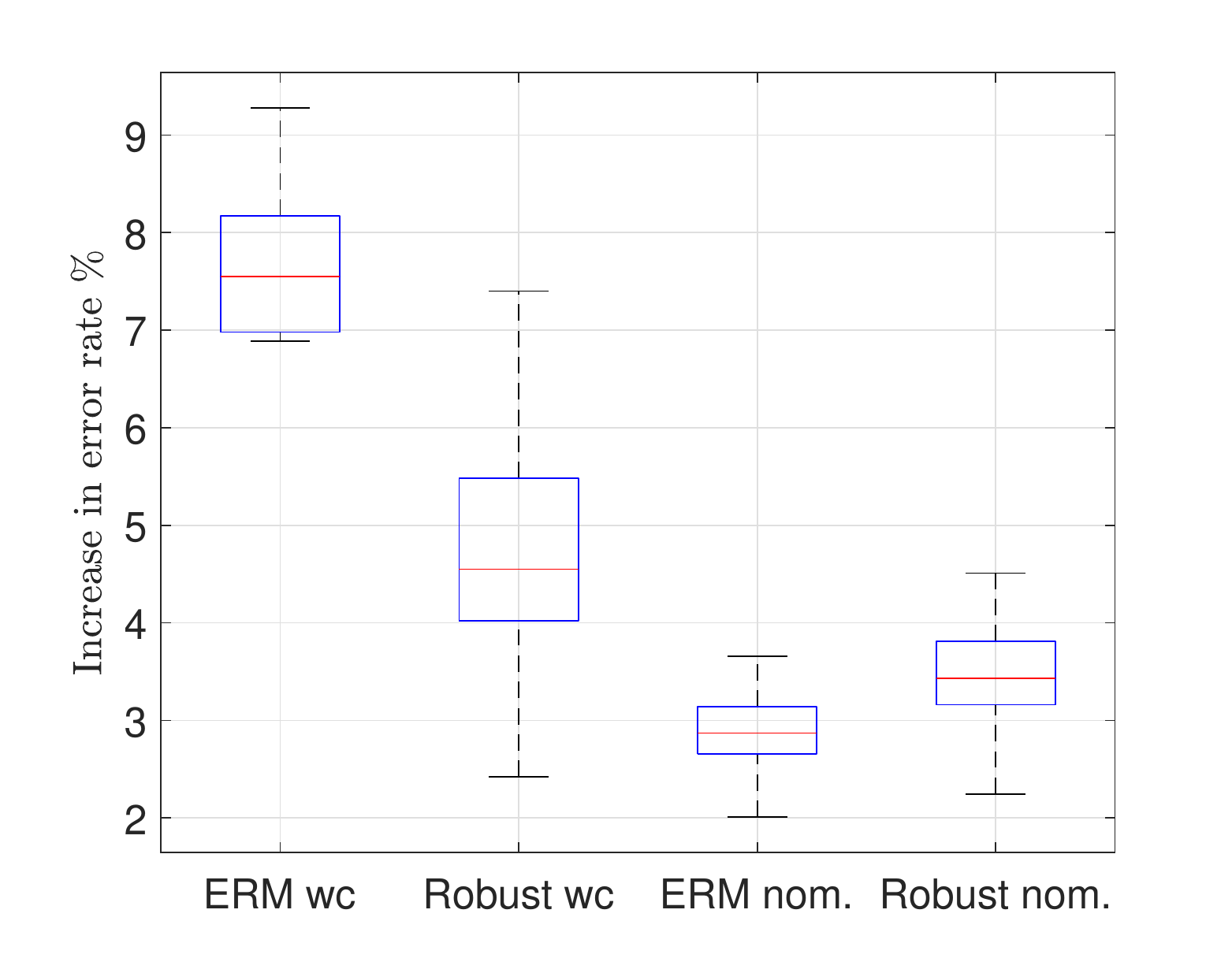}
    \caption{}
    \label{fig:mnist err}
    \end{subfigure}
    \caption{Colored MNIST classification task: (a) Hand-written and colored digits. (b) Box plot of excess error rates under worst-case (wc) and nominal (nominal test conditions, using training data from $50$ Monte Carlo simulations.}
    \label{fig:mnist}
\end{figure*}

\subsection{Colored MNIST classification task} \label{sec:col mnist}

Finally, we consider a binary classification task with real data of handwritten digits \cite{lecun1998gradient} inspired by \cite{arjovsky2019invariant}. The task is to classify images of handwritten digits as \texttt{low} (range $0-4$) and \texttt{high} (range $5-9$), with a corresponding label $y\in \{0,1\}$. For the images we use an autoencoder to extract features $\xvec$ from  grey scale images of size $28 \times 28$. In addition, we consider the images to be (synthetically) colored as \texttt{green} or \texttt{red}, which is indicated by a covariate $x^{\text{col}} \in \{0, 1\}$, see Figure~\ref{fig:mnist digits}. We consider the association between image color and the digits to vary across $|\ctxset| =5$ contexts, i.e.
$$\ptrain(\x^{\text{col}}~=~1|\y~=~0, \ctx)~=~\ptrain(\x^{\text{col}}~=~0|\y~=~1, \ctx) $$
$ \in \{ 0.95, 0.84, 0.72,  0.61, 0.50 \}$ corresponding to $\ctx=1,\dots, 5$. Note that in context $\ctx=5$, the color is entirely uninformative of the digit. We draw $\ndata~=~900$ samples randomly using the MNIST training data with context distribution $\ptrain~=~\{ 0.6, ~0.1, ~0.1, ~0.1, ~0.1\}$. For simplicity, we use a four-dimensional feature vector $\xvec$ since it provides sufficient accuracy in this problem and use a logistic regression model with cross-entropy loss function as in the previous example. The minimum error rates for contexts $1$ to $5$ were estimated as $5.03\%, 14.74\%, 24\%, 31.90\%$ and $34.31\%$, respectively, using $10^4$ data points drawn from the MNIST test data for each context.

We compare our method with only \erm{} here because the numerical solver for minimax proposed by \cite{Sagawa*2020Distributionally} does not provide a reasonable solution with the parameters chosen above. Figure~\ref{fig:mnist err} shows the results using $50$ Monte Carlo simulations. Compared to \erm{}, the robust method once again significantly reduces the excess error rate in the worst case, and results in a minor increase in the nominal case. To obtain $\parmdrr$, we used a step size $\eta~=~0.01$ in Algorithm~\ref{algo: gd danskin}.

\section{Discussion}

We have considered the problem of learning from training data obtained in different contexts. The underlying context distribution is unknown and is estimated empirically, which can underemphasize challenging contexts. Building on the insights from \cite{savage1951theory} and \cite{scarf1957inventory}, we consider all distributions in a \emph{confidence set} and seek the decision parameter that is robust against the distribution that yields the worst \emph{excess} risk. This approach is consistent with the confidence-based  decision-theoretic principles that have recently been proposed  to tackle  imprecise probabilities, cf. \cite{hill2013confidence,bradley2017decision}.

We showed that this approach interpolates between empirical risk minimization and minimax regret objectives based on the chosen confidence level. In this way, it seeks to achieve a statistically motivated trade-off between performance and robustness that is better balanced than the conventional minimax risk approach. Using both real and synthetic data, we demonstrate its ability to provide robustness in worst-case scenarios without harming performance in the nominal scenario. Further work may investigate specific applications in which regularized learning is particularly relevant.

\section*{Acknowledgement}
This research was partially supported by the Swedish Research Council (contract no.: 2018-05040) and the \emph{Wallenberg AI, Autonomous
Systems and Software Program} (WASP) funded by Knut and Alice Wallenberg Foundation.





\end{document}